\pdfoutput=1

\documentclass[11pt]{article}

\usepackage{acl}

\usepackage{times}
\usepackage{latexsym}

\usepackage[T1]{fontenc}

\usepackage[utf8]{inputenc}

\usepackage{microtype}

\usepackage{inconsolata}

\usepackage{pgfplots}
\usepackage[scaled]{helvet}
\usepackage[T1]{fontenc}
\pgfplotsset{compat=newest}
\usepackage{xcolor} 
\usepackage{sansmath} 
\usepackage{subcaption} 
\usepackage{enumitem}

\usepackage{latexsym}
\usepackage{array}
\usepackage{booktabs}
\usepackage{siunitx}
\usepackage{amsmath,amssymb,amsthm}
\usepackage{multirow}

\title{Subword Segmentation for Multilingual Translation: \\Analysing Cross-Lingual Synergy, Interference, and Transfer}
\title{A Systematic Analysis of Subwords in Multilingual Translation}
\title{How Do Subwords Affect Cross-Lingual Tranfer in Multilingual Translation?}
\title{Which Subwords Promote Synergy, Avoid Interference, and Facilitate Cross-lingual Transfer in Multilingual Translation?}
\title{Do Subwords Affect Cross-Lingual Tranfer in Multilingual Translation?}
\title{A Systematic Analysis of Subwords and \\Cross-Lingual Transfer in Multilingual Translation}

\author{Francois Meyer and Jan Buys\\
  Department of Computer Science \\
  University of Cape Town \\
  \texttt{francois.meyer@uct.ac.za, jbuys@cs.uct.ac.za}}

\begin{document}
\maketitle
\begin{abstract}
Multilingual modelling can improve machine translation for low-resource languages, partly through shared subword representations. This paper studies the role of subword segmentation in cross-lingual transfer. We systematically compare the efficacy of several subword methods in promoting synergy and preventing interference across different linguistic typologies. Our findings show that subword regularisation boosts synergy in multilingual modelling, whereas BPE more effectively facilitates transfer during cross-lingual fine-tuning. Notably, our results suggest that differences in orthographic word boundary conventions (the morphological granularity of written words) may impede cross-lingual transfer more significantly than linguistic unrelatedness. Our study confirms that decisions around subword modelling can be key to optimising the benefits of multilingual modelling.

\end{abstract}




\section{Introduction}

Machine translation (MT) models have become increasingly multilingual \citep{10.1145/3406095}. This greatly benefits low-resource languages through positive transfer from high-resource languages \citep{ha-etal-2016-toward, aharoni-etal-2019-massively}. 
However, increasing multilinguality in a limited shared parameter space can lead to suboptimal performance for high-resource languages \citep{firat-etal-2016-multi, nllbteam2022language}. 
There is a tradeoff between maximising positive cross-lingual transfer (also known as synergy) while minimising negative cross-lingual interaction (also known as interference).

Several modelling decisions affect synergy and interference in multilingual MT. \citet{shaham-etal-2023-causes} experimentally analysed the influence of factors like model size and language data proportions. 
One aspect their study failed to consider is subword segmentation.
The shared subword vocabulary of multilingual models presents a similar trade-off as the shared parameter space - overlapping subword representations induce synergy, but having to represent multiple languages in a limited vocabulary can harm cross-lingual transfer \citep{chung-etal-2020-improving, rust-etal-2021-good, patil-etal-2022-overlap}.

In this paper we experimentally analyse the role of subword segmentation in multilingual and cross-lingual MT. Our goal is to compare different classes of subword methods with regards to their ability to induce synergy, reduce interference, and transfer knowledge during cross-lingual finetuning. We also investigate how cross-lingual transfer is influenced by the linguistic similarities of interacting languages, with particular focus on factors related to subword structure like morphological typology and orthographic word boundary conventions (the degree to which morphemes are concatenated or written as separate orthographic words).


\definecolor{customRed}{HTML}{e9b6aa}
\definecolor{customTeal}{HTML}{99a6ad}
\definecolor{customGreen}{HTML}{929e86}
\definecolor{customBlue}{HTML}{e8b961}
\pgfplotsset{ylabel style={yshift=-0.5em},
  tick label style={font=\sansmath\sffamily\scriptsize}, 
  label style={font=\sansmath\sffamily\scriptsize},
  legend style={font=\sansmath\sffamily\scriptsize},
  legend style={draw=none, at={(0.3,1.0)}, anchor=west}, 
      legend columns=4,
  title style={font=\sansmath\sffamily\scriptsize},
  /pgf/number format/.cd,
  use comma,
  1000 sep={},
}
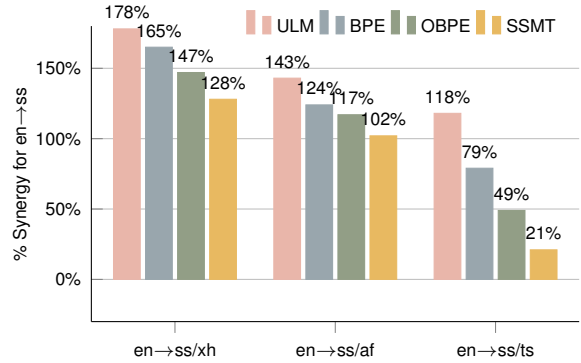
\begin{figure}[t]
  \hspace{-0.5em}
  \begin{tikzpicture}
    \begin{axis}[
      ybar,
      width=8cm,
      height=5.5cm,
      enlarge x limits={abs=1.1cm},
      ylabel={\% Synergy for en$\rightarrow$ss},
      symbolic x coords={en$\rightarrow$ss/xh,en$\rightarrow$ss/af,en$\rightarrow$ss/ts},
      xtick=data,
      nodes near coords={\pgfmathprintnumber\pgfplotspointmeta\%},
      nodes near coords align={vertical},
      nodes near coords style={font=\sansmath\sffamily\scriptsize, text=black},
      axis x line*=bottom,
      axis y line*=left,
      axis line style={-},
      ymin=-30, ymax=180,
      yticklabel={$\pgfmathprintnumber{\tick}\%$},
      ymajorgrids,
      ],
      \addplot[color=customRed, fill=customRed] coordinates {(en$\rightarrow$ss/xh,178) (en$\rightarrow$ss/af,143) (en$\rightarrow$ss/ts,118)};
      \addplot[color=customTeal, fill=customTeal] coordinates {(en$\rightarrow$ss/xh,165) (en$\rightarrow$ss/af,124) (en$\rightarrow$ss/ts,79)};
      \addplot[color=customGreen, fill=customGreen] coordinates {(en$\rightarrow$ss/xh,147) (en$\rightarrow$ss/af,117) (en$\rightarrow$ss/ts,49)};
      \addplot[color=customBlue, fill=customBlue] coordinates {(en$\rightarrow$ss/xh,128) (en$\rightarrow$ss/af,102) (en$\rightarrow$ss/ts,21)};
    \addlegendimage{color=customRed, fill=customRed, ybar, legend image code/.code={\draw[1] (0cm,-0.1cm) rectangle (0.6cm,0.1cm);}}
      \addlegendentry{ULM}
      \addlegendimage{color=customTeal, fill=customTeal, ybar, legend image code/.code={\draw[1] (0cm,-0.1cm) rectangle (0.6cm,0.1cm);}}
      \addlegendentry{BPE}
      \addlegendimage{color=customGreen, fill=customGreen, ybar, legend image code/.code={\draw[1] (0cm,-0.1cm) rectangle (0.6cm,0.1cm);}}
      \addlegendentry{OBPE}
      \addlegendimage{color=customBlue, fill=customBlue, ybar, legend image code/.code={\draw[1] (0cm,-0.1cm) rectangle (0.6cm,0.1cm);}}
      \addlegendentry{SSMT}
    \end{axis}
  \end{tikzpicture}%
  \caption{Performance increase for English$\rightarrow$Siswati through multilingual modelling varies greatly across subword methods and linguistic contexts.}
    \label{synergy_plot}
\end{figure}

\begin{figure*}[h!] 
\small
	\centering
	\begin{tabular}{ll>{\centering\arraybackslash}p{1.8cm}>{\centering\arraybackslash}p{1.9cm}ll} 
		\toprule
		\textbf{Language} & \textbf{Family} & \textbf{Morphology} & \textbf{Orthography} & \textbf{\emph{What is your name?}}  & \textbf{\emph{Thank you!}} \\
		\midrule
		Siswati (ss) & 	
NC/Bantu/Nguni &  agglutinative & conjunctive & Ngubani ligama lakho? & Ngiyabonga! \\
\midrule
            isiXhosa (xh)& NC/Bantu/Nguni &  agglutinative & conjunctive & Ungubani igama lakho? & Enkosi!\\
            Setswana (ts) &  NC/Bantu/Sotho-Tswana  &  agglutinative& disjunctive & Leina la gago ke mang? & Ke a leboga!\\
		Afrikaans (af) & Indo-European/Germanic &  analytic & disjunctive & Wat is jou naam? & Dankie! \\

		\bottomrule
	\end{tabular}
        \captionof{table}{We vary the language modelled alongside Siswati to control relatedness, morphology, and orthography. } 	
 \label{language_details}
\end{figure*}


We run experiments on translation from English to four linguistically diverse South African languages (see Table \ref{language_details}). This selection covers different levels of language relatedness, morphological complexity, and orthographic word granularity, allowing us to analyse how these factors interact with different subword methods to influence cross-lingual transfer. 
Low-resource languages stand to benefit most from multilingual modelling. In all our experiments we focus on cross-lingual transfer to Siswati, which is by far the least resourced among the languages included. It presents exactly the type of real world low-resource translation scenario we are interested in studying.

We conduct two sets of experiments - multilingual MT and cross-lingual finetuning.
Our multilingual experiments follow  \citet{shaham-etal-2023-causes} in training several trilingual MT models and comparing synergy/interference (see Figure \ref{synergy_plot}). In the cross-lingual finetuning experiments we finetune pretrained bilingual MT models on new languages. 
Our results demonstrate that decisions around subword segmentation significantly affect MT performance. 
ULM \citep{kudo-2018-subword} improves synergy in multilingual modelling, while BPE \citep{sennrich-etal-2016-neural} enhances cross-lingual transfer during finetuning.
Going beyond linguistic relatedness, we find that the much less studied influence of orthographic word boundary conventions can drastically affect the cross-lingual transfer achieved between interacting languages. 



\section{Related Work}

Synergy and interference are well-established phenomena \citep{firat-etal-2016-multi, aharoni-etal-2019-massively, nllbteam2022language}, but not well understood. \citet{shaham-etal-2023-causes} address this by systematically analysing the role of several factors in synergy and interference: (1) model size, (2) data size, (3) language proportions, (4) number of languages, and (5) language relatedness. 
Their results show that scaling model size and tuning the data sampling temperature 
greatly alleviates interference.
They do not vary subword segmentation in their experiments, using the same sentencepiece \citep{kudo-richardson-2018-sentencepiece} vocabulary across all models.

However, multilingual vocabularies are known to affect cross-lingual transfer through factors such as cross-lingual subword overlap \citep{pires-etal-2019-multilingual, wu-dredze-2019-beto, patil-etal-2022-overlap} and under-represented low-resource languages \citep{wang-etal-2021-multi-view, acs-2021-exploring}.
These issues have mainly been studied for multilingual language modelling \citep{rust-etal-2021-good, maronikolakis-etal-2021-wine-v, chung-etal-2020-improving}, but the same concerns hold for MT \citep{cho-byte}. We are unaware of existing work comparing the multitude of proposed subword methods in the context of multilingual MT.

\section{Methodology}

This study involves two sets of MT experiments - (1) multilingual (trilingual) experiments to investigate synergy/interference, and (2) finetuning experiments to analyse cross-lingual transfer.
Our goal is to determine which subwords benefit low-resource languages and how cross-lingual transfer depends on linguistic typology.
The linguistic diversity of South Africa is an ideal testing ground for our purposes. 
Siswati is a low-resource agglutinative language, so effective subword modelling is critical for dealing with the inevitably high proportion of out-of-vocabulary words in held-out datasets.
We use Siswati as the low-resource target language in our experiments and alternate the higher resourced language modelled alongside Siswati between isi-Xhosa, Setswana, and Afrikaans. 

Table \ref{language_details} shows how these language present varying linguistic relationships to Siswati. 
IsiXhosa is closely related. 
Setswana is somewhat related and also agglutinative, but diverges in its orthography - its writing system is disjunctive \citep{pretorius-etal-2009-setswana}. This refers to how linguistic words (e.g. nouns, verbs) are represented as orthographic words (space-separated tokens). Disjunctive orthographies write a single linguistic word as multiple orthographic words (e.g. in Setswana prefixal morphemes are space-separeted from verbal roots).

While linguistic relatedness and morphological complexity are obvious features to consider in any analysis of cross-lingual interactions, we are unaware of work considering the impact of orthographic word boundary conventions. We include it as a factor in our study because of its potential relevance to subword segmentation. Orthographic word boundaries determine the pre-tokenisation of text before subword segmenters are applied, so it could well affect aspects like segmentation granularity and overlap between the subword vocabularies of different languages.
Afrikaans is linguistically unrelated to Siswati and also disjunctive, but because of its analytic morphology (lower morpheme-to-word ratio) its written words are sometimes more aligned to those of Siswati (e.g. see phrases in Table \ref{language_details}).

This selection of languages allows us to isolate the cross-lingual effects of linguistic relatedness, morphological typology, and orthographic word boundary conventions. 
In the case of Setswana-Siswati, we can study whether the potentially positive cross-lingual effect of their linguistic relatedness is nullified by the fact that the two languages have very different conventions for orthographic word boundaries.




\subsection{Multilingual Modelling}

We train two bilingual models and one trilingual model per language pair (see Table \ref{multilingual_setup}). 
\begin{figure}[h] 
\small
	\centering
	\begin{tabular}{l>{\centering\arraybackslash}p{1.2cm}l} 
		\toprule
		\textbf{Languages} & \textbf{Examples}  & \textbf{Subwords} \\
  \midrule
		 en$\rightarrow$ss&  166k & BPE/ULM/SSMT\\
             en$\rightarrow$xh/ts/af&  1.6m& BPE/ULM/SSMT\\
            \midrule
            en$\rightarrow$ss+xh/ts/af&  1.6m+166k& BPE/ULM/SSMT/OBPE\\
		\bottomrule
	\end{tabular}
	\captionof{table}{Multilingual experimental setup: bilingual and trilingual models (bilingual OBPE is equivalent to BPE).} 	\label{multilingual_setup}
\end{figure}

\noindent
This setup allows us to compare how MT performance changes for en$\rightarrow$ss and en$\rightarrow$xh/ts/af going from bilingual models to multilingual models. Following \citet{shaham-etal-2023-causes}, we measure synergy/interference for a translation direction $s \rightarrow t$ by the relative difference in performance between a bilingual model trained to translate only from $s$ to $t$ and a multilingual model trained to translate to an additional language. 




\citet{shaham-etal-2023-causes} use test set cross-entropy loss to measure MT performance, but this cannot be reliably used to compare across different subword segmentations. 
Instead, we use test set chrF++ \citep{popovic-2017-chrf} to measure performance. It is a reference-based metric that combines word and character information, so it is well suited for evaluating subword-level performance. Our modified formula for measuring synergy/interference is
\begin{align} \label{interference} 
    \mathcal{I}_{s \rightarrow t} = \frac{
    \mathrm{\textsc{chrF}\texttt{+}\texttt{+}}(M_{s \rightarrow t}^{\mathrm{multi}}) - 
    \mathrm{\textsc{chrF}\texttt{+}\texttt{+}}(M_{s \rightarrow t}^{\mathrm{bi}})}{\mathrm{\textsc{chrF}\texttt{+}\texttt{+}}(M_{s \rightarrow t}^{\mathrm{bi}})}, \nonumber
\end{align}
where $M$ are trained multilingual/bilingual models evaluated on $s \rightarrow t$ translation. Negative values of ${I}_{s \rightarrow t}$ indicate worse performance for $s \rightarrow t$ in the multilingual model (interference) and positive values indicate improved performance (synergy).

\subsection{Cross-Lingual Finetuning}

We train a bilingual subword segmenter and MT model for en$\rightarrow$xh/ts/af, and then finetune and evaluate the model in the other translation directions (e.g. pretrain en$\rightarrow$xh and finetune on en$\rightarrow$ss, en$\rightarrow$ts, and en$\rightarrow$af). Varying the subword method reveals how different subwords facilitate cross-lingual transfer during finetuning from higher resourced languages (isiXhosa/Setswana/Afrikaans) to lower resourced Siswati.

\section{Experimental Setup}


We compare five subword segmenters (four per experiment). We chose methods that represent the main paradigms of subword segmentation - deterministic segmentation, subword regularisation, learning subwords during training, and subword techniques for enhancing cross-lingual transfer.


\noindent \textbf{1. BPE} \citep{sennrich-etal-2016-neural} iteratively adds frequently co-occurring subwords to its vocabulary. We use it as a deterministic segmenter.

\noindent \textbf{2. ULM} \citep{kudo-2018-subword} learns segmentation to optimise a unigram language model and can be used as a probabilistic segmenter, exposing models to multiple subword segmentations for regularisation. We set the sampling parameter $\alpha$ to 0.5. 

\noindent \textbf{3. SSMT} \citep{meyer-buys-2023-subword} is a subword segmental MT model which learns subword segmentation jointly during MT training, with the goal of learning subwords that optimise MT performance.

\noindent \textbf{4. OBPE} \citep{patil-etal-2022-overlap} modifies BPE to boost subword overlap among languages in multilingual  vocabularies. We use it in our multilingual experiments to see if increased shared subword representations promote synergy.

 
\noindent \textbf{5. XBPE} \citep{wang-etal-2020-extending} extends the BPE vocabulary of a pretrained model to include BPE subwords of a new translation direction. New subword embeddings are randomly initialised. We use it in our finetuning experiments to see if the vocabulary extension enhances cross-lingual transfer.

\paragraph{Training}
We train models on WMT22 data \citep{adelani-etal-2022-findings} and validate and test on FLORES \citep{flores, nllbteam2022language}. 
The number of training sentences are shown in Table \ref{multilingual_setup}. 
For en$\rightarrow$ss this is the full WMT22 dataset, but for en$\rightarrow$xh/ts/af we sampled sentences from en$\rightarrow$xh and en$\rightarrow$ts to match the size of en$\rightarrow$af, removing data size as an influence. We also removed examples from en$\rightarrow$xh/ts/af where  English source sentences were found in en$\rightarrow$ss to neutralise the positive transfer effect of multi-parallel overlap \citep{stap-etal-2023-viewing}. The hyperparameters of our models and subword methods are included in Appendix \ref{appendix_modsl}.

\definecolor{customRed}{HTML}{e9b6aa}
\definecolor{customTeal}{HTML}{99a6ad}
\definecolor{customGreen}{HTML}{929e86}
\definecolor{customBlue}{HTML}{e8b961}
\definecolor{newGreen}{HTML}{32a84c}
\definecolor{newOrange}{HTML}{e8c789}
\pgfplotsset{
  tick label style={font=\sansmath\sffamily\scriptsize}, 
  label style={font=\sansmath\sffamily\scriptsize},
  legend style={font=\sansmath\sffamily\scriptsize},
  legend style={draw=none, at={(0.025,1.0)}, anchor=west}, 
      legend columns=4,
  title style={font=\sansmath\sffamily\scriptsize},
  /pgf/number format/.cd,
  use comma,
  1000 sep={},
}

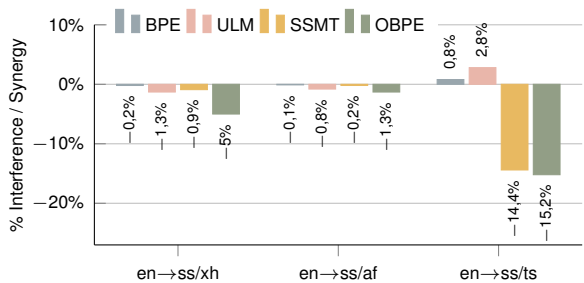
\begin{figure}[t]
  \hspace{-0.5em}
  \begin{tikzpicture}
    \begin{axis}[
      ybar,
      width=8cm,
      height=4.5cm,
      enlarge x limits={abs=1.1cm},
      ylabel={\% Interference / Synergy},
      symbolic x coords={en$\rightarrow$ss/xh,en$\rightarrow$ss/af,en$\rightarrow$ss/ts},
      xtick=data,
      nodes near coords align={horizontal},
      nodes near coords style={font=\sansmath\sffamily\tiny, text=black},
      axis x line*=bottom,
      axis y line*=left,
      axis line style={-},
      ymin=-27, ymax=10,
      yticklabel={$\pgfmathprintnumber{\tick}\%$},
      ymajorgrids,
      every node near coord/.append style={rotate=90,  xshift=0pt, yshift=0pt},
      nodes near coords={
  \pgfmathprintnumber\pgfplotspointmeta\%   
  },
      ],
      \addplot[color=customTeal, fill=customTeal] coordinates {(en$\rightarrow$ss/xh,-0.2) (en$\rightarrow$ss/af,-0.1) (en$\rightarrow$ss/ts,0.8)};
      \addplot[color=customRed, fill=customRed] coordinates {(en$\rightarrow$ss/xh,-1.3) (en$\rightarrow$ss/af,-0.8) (en$\rightarrow$ss/ts,2.8)};
      \addplot[color=customBlue, fill=customBlue] coordinates {(en$\rightarrow$ss/xh,-0.9) (en$\rightarrow$ss/af,-0.2) (en$\rightarrow$ss/ts,-14.4)};
      \addplot[color=customGreen, fill=customGreen] coordinates {(en$\rightarrow$ss/xh,-5.0) (en$\rightarrow$ss/af,-1.3) (en$\rightarrow$ss/ts,-15.2)};

    \addlegendimage{color=customTeal, fill=customTeal, ybar, legend image code/.code={\draw[1] (0cm,-0.1cm) rectangle (0.6cm,0.1cm);}}
      \addlegendentry{BPE}
      \addlegendimage{color=customRed, fill=customRed, ybar, legend image code/.code={\draw[1] (0cm,-0.1cm) rectangle (0.6cm,0.1cm);}}
      \addlegendentry{ULM}
      \addlegendimage{color=customBlue, fill=customBlue, ybar, legend image code/.code={\draw[1] (0cm,-0.1cm) rectangle (0.6cm,0.1cm);}}
      \addlegendentry{SSMT}
      \addlegendimage{color=customGreen, fill=customGreen, ybar, legend image code/.code={\draw[1] (0cm,-0.1cm) rectangle (0.6cm,0.1cm);}}
      \addlegendentry{OBPE}
    \end{axis}
  \end{tikzpicture}%
  \caption{Performance change for en$\rightarrow$xh/af/ts through multilingual modelling alongside en$\rightarrow$ss. }
    \label{interference_plot}
\end{figure}

\begin{figure}[t] 
\small
	\centering
	\begin{tabular}{ll>{\centering\arraybackslash}p{0.8cm}>{\centering\arraybackslash}p{0.8cm}>{\centering\arraybackslash}p{0.8cm}>{\centering\arraybackslash}p{0.8cm}} 
		\toprule
		\textbf{Model} & \textbf{tgt}& \textbf{BPE} & \textbf{ULM} &  \textbf{SSMT} &  \textbf{OBPE} \\
		\midrule
		\multirow{2}{*}{en$\rightarrow$ss/xh} &ss&  33.4	&\textbf{35.1}&	33.6	&31.1  \\
		&xh& 46.8&	\textbf{47.2}	&46.1	&44.6    \\
            \midrule
           \multirow{2}{*}{en$\rightarrow$ss/af} &ss& 28.2&	\textbf{30.6}&	29.7	&27.4  \\
		&af& 60.9	&\textbf{61.1}	&60.2	&60.2    \\
            \midrule
            \multirow{2}{*}{en$\rightarrow$ss/ts} &ss&  22.5	&\textbf{27.5}	&17.8	&18.8  \\
		&ts&  31.3&	\textbf{34.7}	&23.6	&26.4   \\
		\bottomrule
	\end{tabular}
	\captionof{table}{Test set chrF++ of trilingual models.} 	\label{trilingual}
\end{figure}

\section{Results \& Discussion}

We plot the synergy/interference analysis of our multilingual experiments in Figures \ref{synergy_plot} \& \ref{interference_plot}, while the absolute performance of the models are provided in Table \ref{trilingual}. The results from our cross-lingual finetuning experiments are presented in Figure \ref{heatmap}.

\subsection{Which subwords promote synergy and minimise interference?}

ULM consistently achieves greater synergy than other subword methods. This holds across all linguistic contexts (Fig. \ref{synergy_plot}) and results in better absolute performance in all translation directions (Table \ref{trilingual}). It comes at the cost of minimal interference for the higher resourced languages, and even some synergy for en$\rightarrow$ts (Fig. \ref{interference_plot}). The subword regularisation of ULM ensures that models are more robust to the varied subwords of multilingual modelling.

\subsection{Which subwords transfer cross-lingually?}

BPE subwords exhibit the greatest cross-lingual transferability.
In contrast to our multilingual findings, the subword regularisation of ULM proves a barrier to cross-lingual finetuning. ULM is a probabilistic segmenter that is sampled during training, but when the probabilistic model is based on one language and applied to another,
its samples might yield highly inadequate subword units.  
The consistent deterministic segmentation of BPE allows the finetuned model to adapt to a new translation direction effectively. 

\begin{figure}[t] 
  \centering

  {\small Finetuned and evaluated on en $\rightarrow$}

  \hspace{-0.1cm}
  \rotatebox[origin=c]{90}{\parbox{0.1\textheight}{\small \hspace{-0.8cm} Pretrained on en $\rightarrow$}}\hspace{-0.1cm}
  \captionsetup[subfigure]{oneside,margin={0.8cm,0cm}}
  \begin{subfigure}[t]{0.48\columnwidth}
  \centering
  \begin{tikzpicture}[scale=0.6]
  \foreach \y [count=\n] in {
      {46.9,60.4,38.7,32.3},
    {46.4,61.0,36.9,28.3},
    {43.8,58.5,31.1,25.5},
    } {
       \foreach \a [count=\i] in {xh,af,ts,ss} {
        \node[font=\mdseries\sansmath\sffamily\small] at (1.35*\i,0) {\a};
      }
      \foreach \x [count=\m] in \y {
        \node[fill=newGreen!\x!newOrange, minimum width=8.1mm, minimum height=6mm, text=white,font=\sansmath\sffamily\small] at (1.35*\m,-\n) {\x};
      }
    }

  \foreach \a [count=\i] in {xh,af,ts} {
    \node[font=\sansmath\sffamily\small] at (0,-\i) {\a};
  }
\end{tikzpicture}
\subcaption{BPE}
\end{subfigure}
\captionsetup[subfigure]{oneside,margin={-0.2cm,0cm}}
 \begin{subfigure}[t]{0.48\columnwidth}
 \centering
  \begin{tikzpicture}[scale=0.6]
  \foreach \y [count=\n] in {
      {47.8,15.8,1.9,31.1},
      {13.5,61.6,17.3,28.2},
      {10.8,3.6,33.8,24.7},
    } {
       \foreach \a [count=\i] in {xh,af,ts,ss}  {
        \node[font=\mdseries\sansmath\sffamily\small] at (1.35*\i,0) {\a};
      }
      \foreach \x [count=\m] in \y {
        \node[fill=newGreen!\x!newOrange, minimum width=8.1mm, minimum height=6mm, text=white,font=\sansmath\sffamily\small] at (1.35*\m,-\n) {\x};
      }
    }

\end{tikzpicture}
  \subcaption{ULM}
  \end{subfigure}

    \vspace{-0.5cm}
  \hspace{0.1cm}
  \captionsetup[subfigure]{oneside,margin={0.8cm,0cm}}
\begin{subfigure}[t]{0.48\columnwidth}
  \centering
  \begin{tikzpicture}[scale=0.6]
  \foreach \y [count=\n] in {
      {46.5,59.9,38.0,32.2},
    {46.6,60.3,36.3,30.1},
    {42.0,56.8,27.6,23.4},
    } {
       \foreach \a [count=\i] in {xh,af,ts,ss}  {
        \node[font=\mdseries\sansmath\sffamily\small] at (1.35*\i,0) {\a};
      }
      \foreach \x [count=\m] in \y {
        \node[fill=newGreen!\x!newOrange, minimum width=8.1mm, minimum height=6mm, text=white,font=\sansmath\sffamily\small] at (1.35*\m,-\n) {\x};
      }
    }

  \foreach \a [count=\i] in {xh,af,ts} {
    \node[font=\sansmath\sffamily\small] at (0,-\i) {\a};
  }
\end{tikzpicture}
\subcaption{SSMT}
\end{subfigure}
\captionsetup[subfigure]{oneside,margin={-0.2cm,0cm}}
 \begin{subfigure}[t]{0.48\columnwidth}
 \centering
  \begin{tikzpicture}[scale=0.6]
  \foreach \y [count=\n] in {
      {46.9,59.5,37.3,29.9},
    {46.1,61.0,35.6,25.0},
    {43.4,58.3,31.1,21.9},
    } {
       \foreach \a [count=\i] in {xh,af,ts,ss}  {
        \node[font=\mdseries\sansmath\sffamily\small] at (1.35*\i,0) {\a};
      }
      \foreach \x [count=\m] in \y {
        \node[fill=newGreen!\x!newOrange, minimum width=8.1mm, minimum height=6mm, text=white,font=\sansmath\sffamily\small] at (1.35*\m,-\n) {\x};
      }
    }

\end{tikzpicture}
  \subcaption{XBPE}
  \end{subfigure}
  \caption{Test set chrF++ of pretraining for en$\rightarrow$xh/af/ts (rows) and finetuning on en$\rightarrow$xh/af/ts/ss (columns). Diagonal entries are bilingual models with no finetuning. }
  \label{heatmap}
\end{figure}

\subsection{What is the role of linguistic typology?}

A consistent pattern emerges in the cross-lingual dynamics between Siswati and other languages. 
IsiXhosa modelling proves to be most beneficial for Siswati performance.
Afrikaans achieves less transfer, presumably because it is not related. 
Somewhat surprisingly, the weakest synergy is between Siswati and Setswana, even though both are agglutinative Bantu languages.
This highlights the impact of orthographic systems on cross-lingual transfer: \emph{diverging word boundary conventions can impede cross-lingual transfer more than linguistic unrelatedness}. 
Data-driven multilingual models that learn from text might miss underlying similarities between languages that are obscured by superficial differences in their surface realisations.

Our results highlight two interacting effects. 
Firstly, linguistic relatedness does play a role -- isiXhosa consistently improves Siswati more than Setswana and Afrikaans. 
Secondly, in the specific case of Setswana-Siswati, their relatedness is rendered all but irrelevant by the fact that the languages have diverging orthographies.
Afrikaans does not have the extremely disjunctive orthography of Setswana so even though it is less related to Siswati than Setswana, the orthography of Setswana prevents transfer to Siswati. 
Linguistic distance plays a role in both cases (Afrikaans-Siswati and Setswana-Siswati) but for Setswana-Siswati it is a less important factor than orthography.

Orthography is a notable difference between Nguni languages like Siswati and Sotho-Tswana languages like Setswana \citep{pretorius-etal-2009-setswana}.
\citet{taljard-bosch-2006-comparison} showed that the diverging orthographies of these two language groups necessitate  different approaches to more traditional NLP tasks, even though the languages are linguistically and morphologically related.
Our results suggest a similar situation for cross-lingual transfer in multilingual modelling: Differences in orthographic word boundary conventions harms synergy between otherwise related languages.

\section{Conclusion}

We presented an in-depth study on the role of subwords in multilingual and cross-lingual MT. 
Our results demonstrate that subword segmentation significantly influences cross-lingual interactions.
ULM proves optimal for transfer to low-resource languages in multilingual modelling, while BPE enables greater cross-lingual transfer during finetuning.
Besides language relatedness, we show that similarities/differences in orthographic word granularity can greatly affect multilingual performance.
There is more work to be done on the role of orthographic word boundary conventions in neural MT. 
Future work could aim to design multilingual techniques 
that see past orthographic differences in order to leverage more fundamental similarities between languages.


\section*{Limitations}

Our study is limited to translation from English to four South African languages. While the chosen languages are typologically diverse, our conclusions might not necessarily hold for languages from different language families and with distinct orthographies. We did not consider languages that have multiple orthographies, which might be another approach to study the effects of orthography. 
The performance differences between different subword segmentation methods across languages in our results are relatively consistent, but a more detailed analysis on the interaction between choice of subword segmentation method and language could yield additional explanations of the results. 

\section*{Acknowledgements}
This work is based on research supported in part by the National Research Foundation of South Africa (Grant Number: 129850).  
Computations were performed using facilities provided by the University of Cape Town’s ICTS High Performance Computing team: \url{hpc.uct.ac.za}.

\appendix

\section{Model Configurations}
\label{appendix_modsl}

We use the model size and training hyperparameters of the fairseq \href{https://github.com/facebookresearch/fairseq/blob/main/fairseq/models/transformer/transformer_legacy.py}{transformer-base}  architecture. We train our models for 45 epochs and use a language sampling temperature of $T=1.5$ to balance exposure to low-resource and high-resource languages. In our cross-lingual finetuning experiments we finetune models on en$\rightarrow$ss for 20 epochs and on en$\rightarrow$xh/af/ts for 10 epochs. 

When applying subword methods we specify a shared vocabulary size of 8k. This is slightly smaller than the optimal vocabulary size of 10k used by \citet{meyer-buys-2023-subword} in experiments on these same languages, since we use smaller subsets of the WMT22 datasets.

\end{document}